\documentclass{article}

\usepackage{PRIMEarxiv}

\usepackage[utf8]{inputenc} 
\usepackage[T1]{fontenc}    
\usepackage{hyperref}       
\usepackage{url}            
\usepackage{booktabs}       
\usepackage{amsfonts}       
\usepackage{nicefrac}       
\usepackage{microtype}      
\usepackage{lipsum}
\usepackage{natbib}
\usepackage{fancyhdr}       
\usepackage{graphicx}       
\graphicspath{{media/}}     

\title{On Feature Scaling of Recursive Feature Machines}

\author{
  Arunav Gupta, Rohit Mishra, William Luu, Mehdi Bouassami \\
  Halıcıoğlu Data Science Institute \\
  UCSD \\
  \texttt{\{arg002, r1mishra, wjluu, mbouassa\}@ucsd.edu} \\
}

\begin{document}
\maketitle

\begin{abstract}
In this technical report, we explore the behavior of Recursive Feature Machines (RFMs), a type of novel kernel machine that recursively learns features via the average gradient outer product, through a series of experiments on regression datasets. When successively adding random noise features to a dataset, we observe intriguing patterns in the Mean Squared Error (MSE) curves with the test MSE exhibiting a decrease-increase-decrease pattern. This behavior is consistent across different dataset sizes, noise parameters, and target functions. Interestingly, the observed MSE curves show similarities to the "double descent" phenomenon observed in deep neural networks, hinting at new connection between RFMs and neural network behavior. This report lays the groundwork for future research into this peculiar behavior. 
\end{abstract}


\section{Introduction}

Recent work into understanding the theory behind recent advances in the breakthrough performance of large neural networks \citep{GPT3,DALLE,Midjourney,llama} has analyzed kernel machines as more theoretically accessible approximations of neural networks \citep{belkin2018understand,belkin2019reconciling,FeatureLearningEmergenceShi} The Recursive Feature Machine, as introduced in \cite{radhakrishnan23} describes a new type of kernel machine which recursively learns a Mahalanobis norm \citep{MahanolobisDistance} using the average gradient outer product. The original paper indicates several theoretical and empirical results which connect the RFM to the Neural Tangent Kernel \citep{NTK}, as well as commonly observed behavior in neural networks such as the arrangement of weights in a dense layer of a deep neural network. 

In this technical report, we describe a series of experiments which show interesting behavior under feature scaling for RFMs. As we add random noise features to a regression dataset, MSE after training appears to decrease, then increase, then decrease again. Results are compared to "baseline" kernels which use the Laplacian norm (these are also simply RFM kernels that have been trained for 0 iterations).

Some discussion on the results is provided, as the curves appear to look similar to "double descent" as seen in deep neural networks. While random feature-style theoretical setups have been studied in the context of kernel machines \citep{RR08,mei2020generalization}, more work in this direction is required, and this report hopes to serve as the baseline for further research into this peculiar behavior.

\section{Experiment Setup}
The RFM is trained according to the algorithm described in \citep{radhakrishnan23}. The models are trained at full precision on a single NVIDIA RTX 2060 6GB VRAM running Ubuntu 20.04 with Pytorch v1.13 \citep{Paszke2019} for 10 iterations. At the beginning of training, 20\% of the training set is reserved for validation; after each iteration, MSE is measured on the validation set and the best $M$ is kept. Empirically, the best model is found around the 3rd or 4th iteration, and in rare cases the validation MSE improves marginally from the 3rd iteration to the 10th iteration.

For the base experiment, a dataset of size $1000 \times 2000$ (called $X$) is randomly generated, with each value being i.i.d from $\mathcal{N}(0, 1)$. Variations to the base experiment are described in the relevant sections below.

For every $d \in [5, 6, 7, \dots, 99] \cup [100, 110, 120, \dots, 2000]$, we take the first $d$ columns  of $X$ and divide them by $\frac{1}{\sqrt{d}}$ in order to control the standard deviation. A 80-20 train-test split is applied, and target values are generated using a simple cubic function:
\begin{equation} \label{eqn:cubic}
    f(x) = 5x_1^3 + 2x_2^2 + 10x_3
\end{equation}
Note that the target equation \ref{eqn:cubic} only utilizes the first three values in each $x$. This means, for each $d$, there are $d - 3$ columns of pure noise appended, which do not have any contribution to the target value. Thus, every model is given all data required to compute the exact value of the target — all fluctuations in model performance are due to the model's ability to learn the target function.

Finally, the RFM is trained for 10 iterations with cross-validation, and the MSE is computed for the train and test splits.

The full scaling test is repeated 100 times with different randomly generated $X$ in order to further validate the results.
\section{Results}
\begin{figure}[!htb]
    \centering
    \includegraphics[scale=0.5]{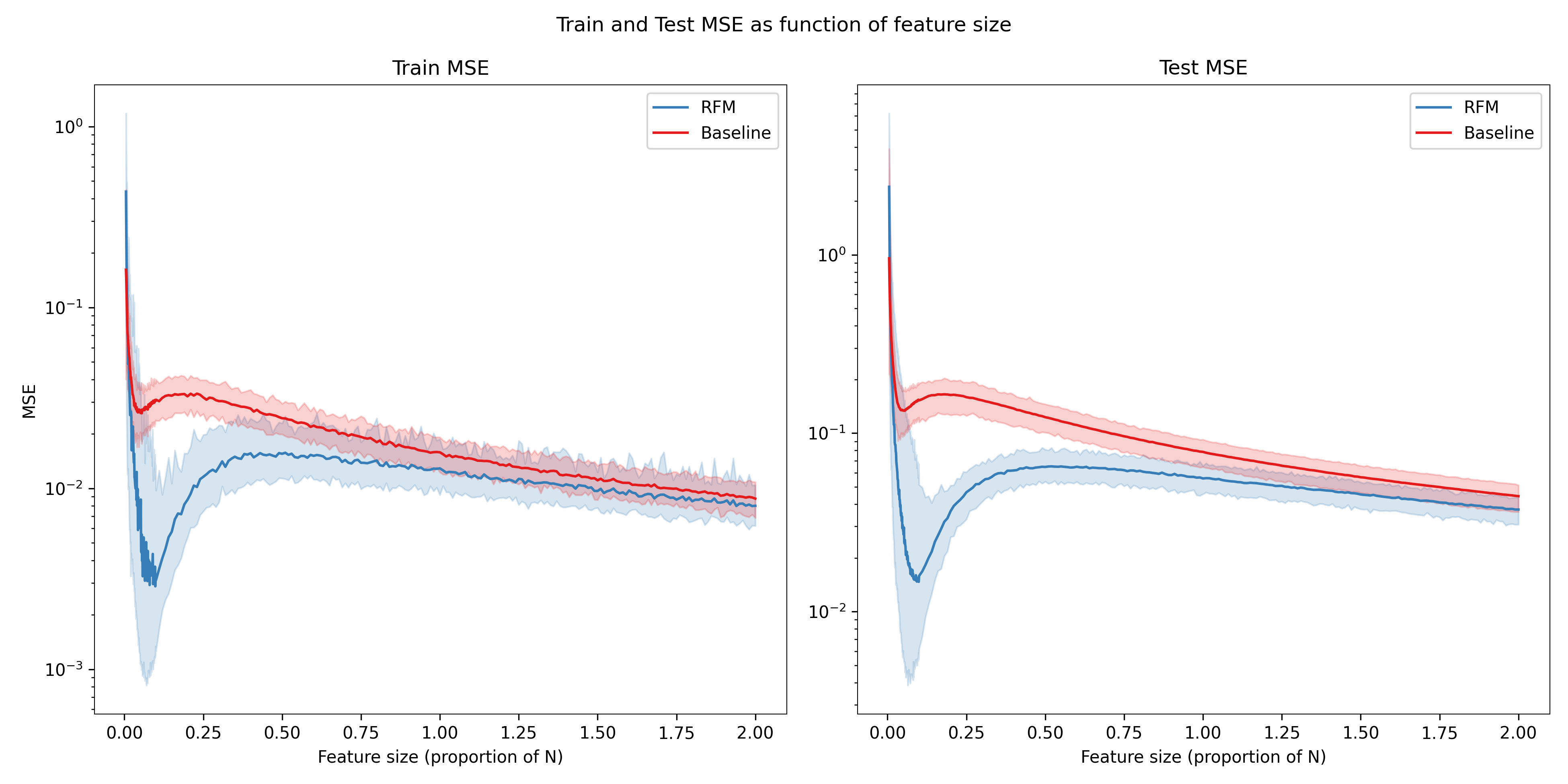}
    \caption{Train and Test MSE as a function of feature size for RFM and Baseline (Laplacian) kernel. 95\% confidence intervals are shown by the respectively colored shaded areas.}
    \label{fig:base}
\end{figure}

Figure \ref{fig:base} shows the result for the base experiment with $(N, D) = (1000, 2000)$. Results of variations are described in the subsequent sections below.
\subsection{Noise Effects}
\begin{figure}[!htb]
    \centering
    \includegraphics[scale=0.5]{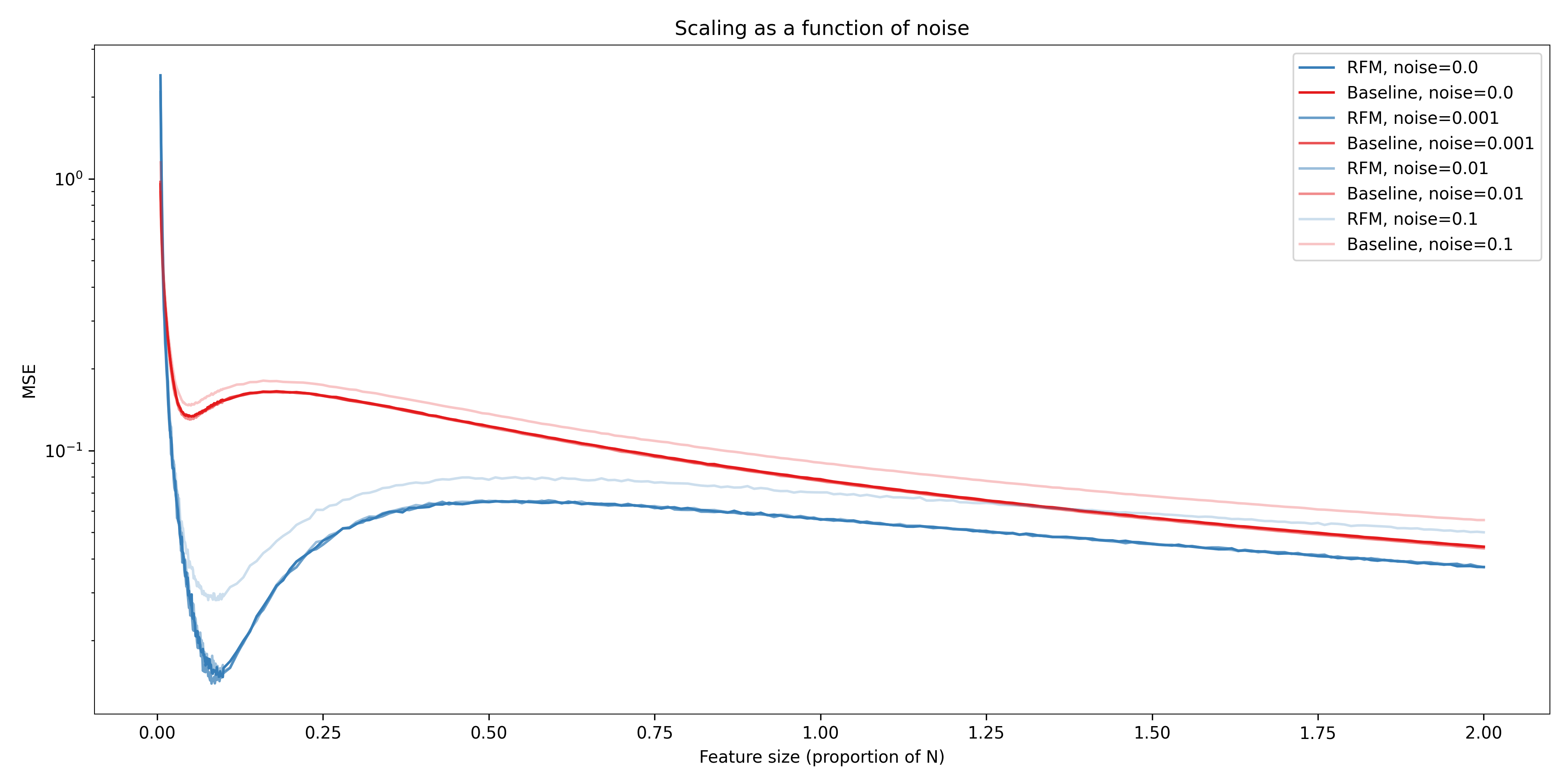}
    \caption{Noise effects, from  $\sigma = 0.0$ to $\sigma = 0.1$. Red is baseline kernel, blue is RFM.}
    \label{fig:noise}
\end{figure}

In order to better understand how dataset noise affects the RFM's performance, we successively add higher amounts of noise to the target function. Random gaussian noise was added to the target function like so:
\begin{equation}
    y = f(x) + \mathcal{N}(0, \sigma)
\end{equation}
Where $\sigma \in [0, 0.1, 0.01, 0.001]$ are the tested noise values. For consistency, the same random noise is used for each tested $d$ value. Figure \ref{fig:noise} shows the results of the noise experiment.
\subsection{Dataset Size Effects}
\begin{figure}[!htb]
    \centering
    \includegraphics[scale=0.5]{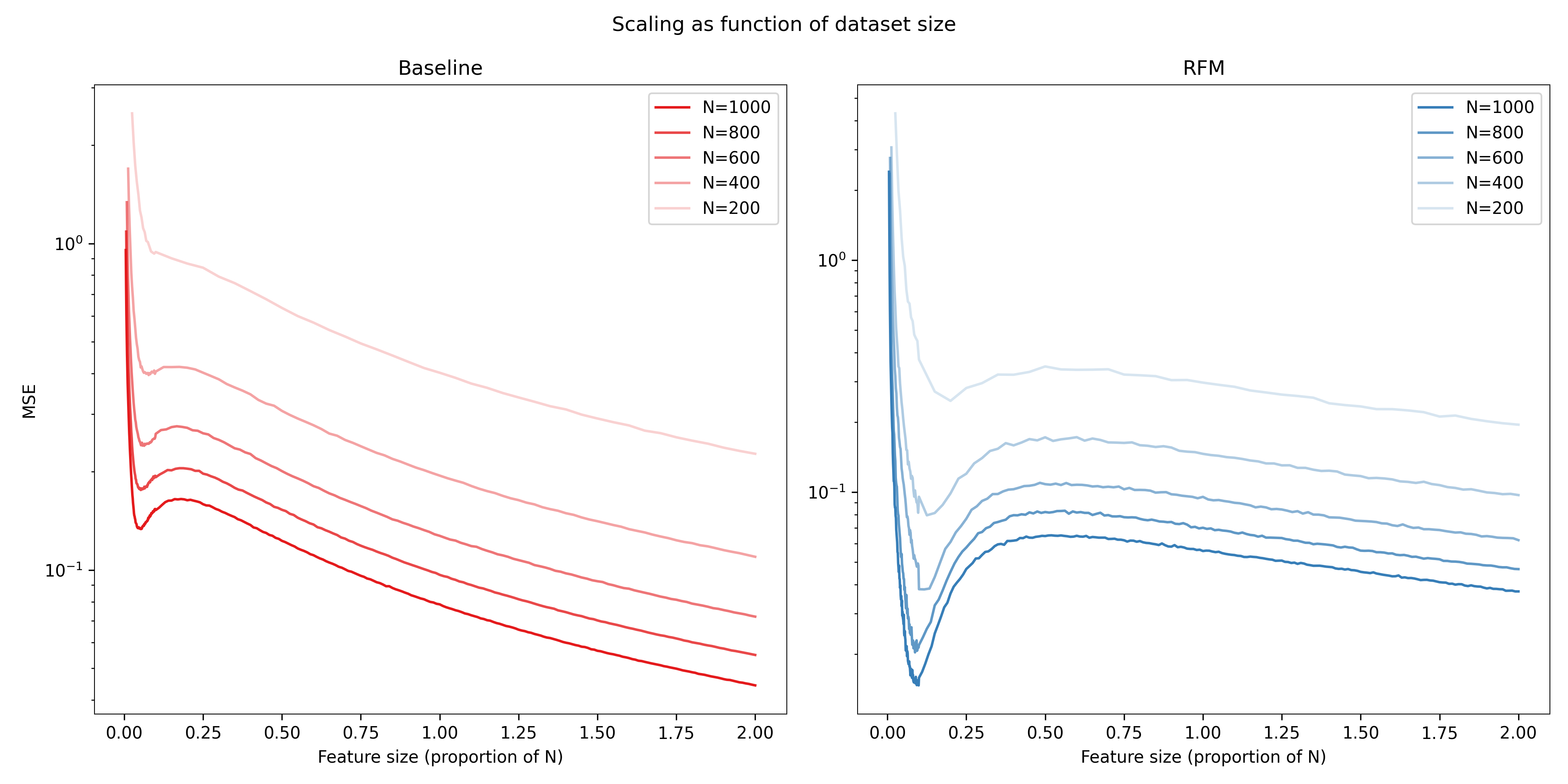}
    \caption{Effects of dataset size, from $N = 200$ to $N = 1000$. Feature size is scaled from $5$ to $2N$ for each dataset. Left (red) is baseline, right (blue) is RFM.}
    \label{fig:size}
\end{figure}

We also test if dataset size has any affect on the scaling behavior. Dataset sizes of $N \in [200, 400, 600, 800, 1000]$ are tested, and for each, the values of $d$ are scaled proportionally: $d = \left[5, 6, 7, \dots, \frac{N}{10}\right] \cup \left[\frac{N}{10}, \frac{N}{10} + 10, \frac{N}{10} + 20, \dots, 2N\right]$. Figure \ref{fig:size} shows the results of the dataset size experiment.
\subsection{Target Function Effects}
\begin{figure}[!ht]
    \centering
    \includegraphics[scale=0.5]{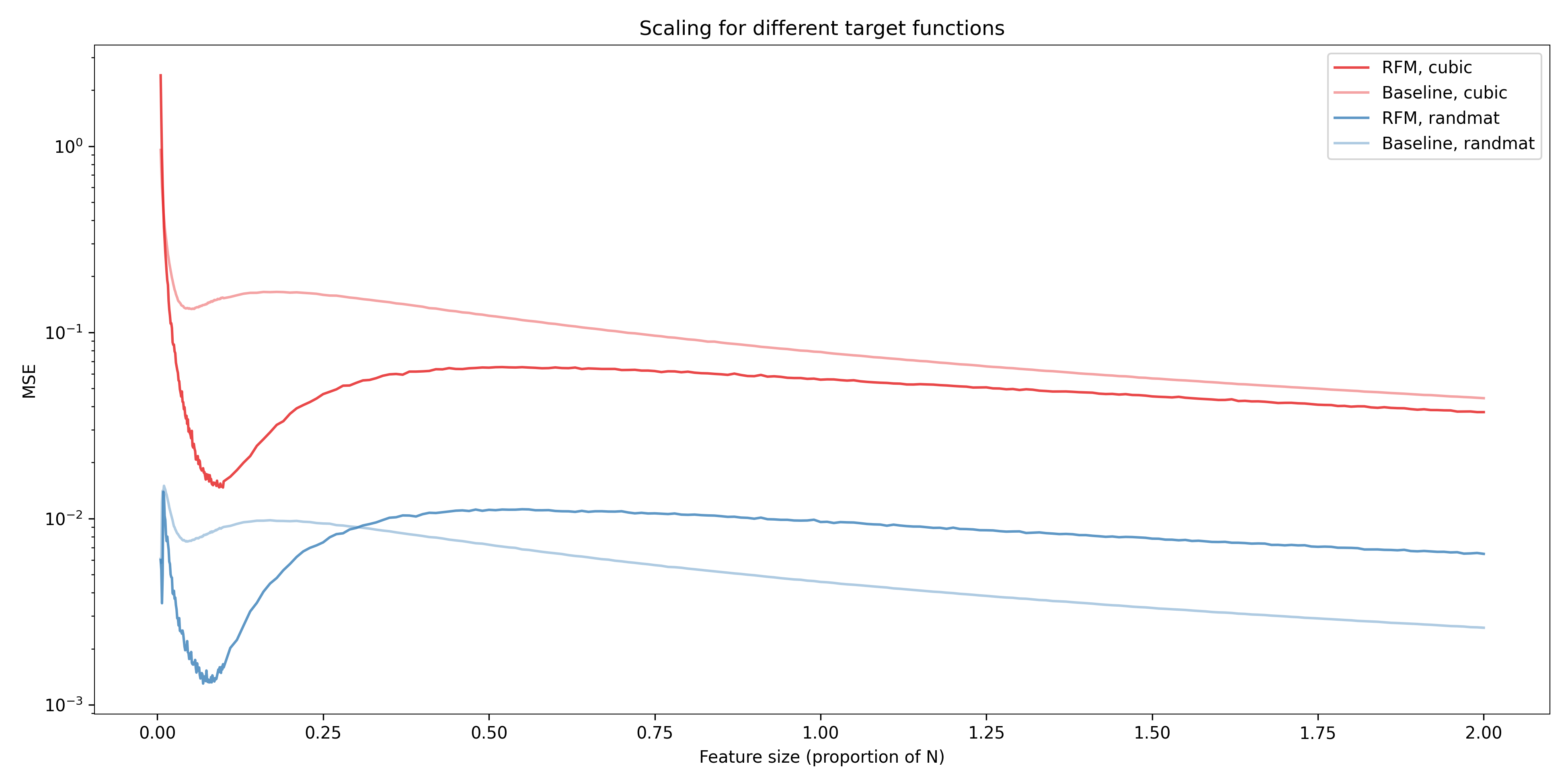}
    \caption{Scaling for cubic and random matrix functions, $(N, D) = (1000, [5,2000])$.}
    \label{fig:function}
\end{figure}

Our final experiment tests a different target function and compares the result to the original target function defined in equation \ref{eqn:cubic} (labeled as "cubic" on figure \ref{fig:function}). The alternative target function, called "randmat" is defined by equation \ref{eqn:randmat}:
\begin{equation}\label{eqn:randmat}
    f(X) = X'K \quad\quad \texttt{where}~ K \sim \mathcal{N}_{10 \times 1}(0,1)
\end{equation}
And $X'$ is the first 10 columns of $X$. Figure \ref{fig:function} compares the results of the alternative target function to the cubic target function.
\section{Conclusion}
Overall, the results indicate that despite modifications to the dataset size, noise parameter, and target function, the shape of the test MSE curve as a function of feature size does not change. In all cases, the RFM MSE drops sharply until $D = 0.1N$, then rises until around $D = 0.5N$, and then continues a shallower descent back down until the end of the experiment at $D = 2N$. Interestingly, the Laplacian kernel also follows this pattern, albeit with a much shorter turnaround (the inflection points occur at around $D = 0.05N$ and $D = 0.2N$, respectively).

The type of dataset and target function being used in these experiments make the shapes the MSE follow highly peculiar. For one, only the first three variables of each datapoint are used to compute the target value, so as additional "noise" features are added, the MSE is not expected to drop (thus indicating a performance increase) in the region to the left of $D = 0.1N$. We can confirm that the values in $X$ are indeed i.i.d — multicollinearity is centered at zero across all features. More perplexing, however, is the region to the right of $D = 0.5N$, where the test MSE begins to fall slowly. As this happens in all cases, it appears to be an artifact of the model, not the dataset.

The authors note that these MSE curves look similar to "double descent" curves seen in deep neural networks \citep{Nak19}. However, more analysis is needed before confirming this relationship. Indeed, early evidence points in this direction, as the feature size of the RFM directly relates to the size of the $M$-norm matrix (which is $D\times D$). The authors of the original paper theorize a relation between the $M$-norm matrix and the first layer weights of a deep fully connected neural network. Thus, it is possible that increasing the feature size by adding random noise features, we are performing an operation analogous to increasing the width of a neural network, thus inducing a double-descent-style phenomenon.

All experiments shown in this report are reproducible by using the code at \footnote{\href{https://github.com/agupta01/ml-theory-capstone}{https://github.com/agupta01/ml-theory-capstone}} and following the instructions under the `Scaling Test` section in the README.
\section*{Acknowledgments}
Experiments described in this technical report were done under the HDSI DSC 180 Capstone program. The authors are immensely grateful to Mikhail Belkin and Parthe Pandit from UCSD for their mentorship and guidance during the project.

\bibliography{references}

\end{document}